%% file: main_arxiv.tex
\pdfoutput=1
\documentclass{article}

\usepackage[whole]{bxcjkjatype}
\usepackage[T1]{fontenc}
\usepackage{lmodern}
\usepackage{amsmath,amssymb}

\usepackage[a4paper,margin=1in]{geometry}  
\usepackage{setspace}  

\usepackage{titlesec}  
\titleformat{\section}{\large\bfseries}{\thesection}{1em}{}

\usepackage{amsmath,amssymb,bm}  

\usepackage{graphicx}  
\usepackage{caption}  
\usepackage{natbib}  
\usepackage[colorlinks=true, linkcolor=blue, citecolor=blue, urlcolor=blue]{hyperref}

\usepackage[whole]{bxcjkjatype}
\usepackage{amsmath,amssymb,bm}
\usepackage{cleveref} 
\usepackage{xspace}

\usepackage{booktabs}
\usepackage{makecell}
\usepackage[separate-uncertainty]{siunitx}
\input{preemble/01_my_packages}
\input{preemble/02_math_commands}

\input{preemble/03_my_commands}

\setboolean{isdoublecolumn}{false}

\title{%
\huge
Syntactic Learnability of \\
Echo State Neural Language Models at Scale}
\author{%
Ryo Ueda$^{1}$ \quad
Tatsuki Kuribayashi$^{2}$ \quad
Shunsuke Kando$^{1}$ \quad
Kentaro Inui$^{2,3,4}$ \\[2ex]
$^{1}$The University of Tokyo \quad
$^{2}$MBZUAI \quad
$^{3}$Tohoku University \quad
$^{4}$RIKEN \\[2ex]
\texttt{\{ryoryoueda, skando\}@is.s.u-tokyo.ac.jp} \\
\texttt{\{Tatsuki.Kuribayashi, Kentaro.Inui\}@mbzuai.ac.ae}
}

\date{}

\begin{document}

\maketitle

\begin{abstract}
\input{section/00_abstract}
\end{abstract}
\input{section/01_introduction}
\input{section/02_esn}
\input{section/03_why_esn}
\input{section/04_experiments}

\input{section/05_conclusion}

\section*{Acknowledgments}
This work was supported by JSPS KAKENHI (Grant Number JP23KJ0768) and JST ACT-X (Grant Number JPMJAX24C5).

\bibliography{%
reference/emcom,%
reference/esn,%
reference/linguistics,%
reference/ml}

\newpage

\appendix

\input{appendix/a_connectivity}

\newpage

\input{appendix/b_leaking_rate}

\end{document}

%% file: preemble/01_my_packages.tex
\usepackage{ifthen}
\usepackage{autobreak}

%% file: preemble/02_math_commands.tex
\def\eqref#1{equation~\ref{#1}}

\def\1{\bm{1}}

\def\va{{\bm{a}}}
\def\vb{{\bm{b}}}

\def\vh{{\bm{h}}}

\def\vo{{\bm{o}}}

\def\vu{{\bm{u}}}
\def\vv{{\bm{v}}}

\def\mA{{\bm{A}}}
\def\mB{{\bm{B}}}

\def\mM{{\bm{M}}}

\def\mV{{\bm{V}}}
\def\mW{{\bm{W}}}

\DeclareMathAlphabet{\mathsfit}{\encodingdefault}{\sfdefault}{m}{sl}
\SetMathAlphabet{\mathsfit}{bold}{\encodingdefault}{\sfdefault}{bx}{n}

\newcommand{\R}{\mathbb{R}}

\newcommand{\softmax}{\mathrm{softmax}}



%% file: preemble/03_my_commands.tex
\newcommand{\stateSize}{N_{\textup{state}}}
\newcommand{\vocabSize}{N_{\textup{vocab}}}

\newcommand{\hpSpectralRadius}{\rho_{\textup{rec}}}  
\newcommand{\hpInputScale}{\sigma_{\textup{in}}}     
\newcommand{\hpMinLeakingRate}{\alpha_{\textup{min}}}
\newcommand{\hpMaxLeakingRate}{\alpha_{\textup{max}}}
\newcommand{\hpConnectivity}{\gamma}  
\newcommand{\hpActivation}{f} 
\newcommand{\hpOutputRank}{r_{\textup{out}}}  
\newcommand{\hpRecDegree}{d}  

\newcommand{\mRec}{\mW_{\textup{rec}}}
\newcommand{\mInput}{\mW_{\textup{in}}}
\newcommand{\mOutput}{\mW_{\textup{out}}}
\newcommand{\vBias}{\vb_{\textup{out}}}
\newcommand{\vLeaking}{\va}

\newcommand{\iidsample}{\mathrel{\overset{\textup{i.i.d}}{\sim}}}

\newcommand{\trainableParam}{\bm{\theta}_{\textup{Tr}}}
\newcommand{\frozenParam}{\bm{\theta}_{\textup{Fr}}}
\newcommand{\totalParam}{\bm{\theta}_{\textup{All}}}

\newcommand{\Transformer}{Transformer\xspace}
\newcommand{\LSTM}{LSTM\xspace}
\newcommand{\ESN}{ESN\xspace}

\newboolean{isdoublecolumn}

%% file: section/00_abstract.tex
What is a neural model with minimum architectural complexity that exhibits reasonable language learning capability? 
To explore such a simple but sufficient neural language model, we revisit a basic reservoir computing (RC) model, Echo State Network (ESN), a restricted class of simple Recurrent Neural Networks.
Our experiments showed that ESN with a large hidden state is comparable or superior to Transformer in grammaticality judgment tasks when trained with about 100M words, suggesting that architectures as complex as that of Transformer may not always be necessary for syntactic learning.

%% file: section/01_introduction.tex
\section{Introduction}
\label{section.introduction}

If there exists a neural language model (LM) that has an architecture with minimum complexity yet has a language acquisition capability comparable to humans, what would it look like?
What architecture would meet such a minimum requirement?
To answer such scientific questions ultimately, we begin with a neural LM that is as simple as possible and evaluate its language acquisition capability, especially on syntactic generalization.
Such a question can not directly be answered with the recent success of Transformer-based large LMs (LLMs)~\citep{VaswaniSPUJGKP-2017-transformer} alone, where the settings/architectures might be overly complex beyond cognitively-inspired motivations.

In this paper, we adopt a neural network class called \emph{Echo State Network} \citep[\ESN,][]{Jaeger-2001-ESN, LukoseviciusJ-2009-reservoir, RodanT-2011-minimum-complexity-esn, Lukosevicius-2012-practical}.
It is a special class of simple Recurrent Neural Networks in which only an output matrix is trainable, while the other parameters, e.g., recurrent matrix, are fixed (frozen) after initialization.
Despite its highly restrictive nature, \ESN has widely been used in time series processing and has demonstrated surprising capability \citep{TanakaYHNKTNNH-2019-recent}; for example, \ESN served as a strong baseline for text classification tasks \citep{WietingK-2019-no, CabessaHKLL-2021-efficient-text-classification} as well as a subject of analyses in computational (psycho-)linguistic studies~\citep{TongBCC-2007-gammatical-esn, FrankC-2008-esn, FrankB-2011-insensitivity}.

In our experiments, we trained ESN-based LMs with about 100M words, to which human children are thought to be exposed by 13 years old, at a larger scale than the age ESN being actively explored~\citep{TongBCC-2007-gammatical-esn}.
We compared them with \LSTM \citep{HochreiterS-1997-lstm} and \Transformer LMs.
Our experimental results demonstrate that ESN with a sizeable hidden state was comparable or superior to \Transformer trained from scratch, while \LSTM showed the best result in our setting, suggesting the benefits of gate mechanisms in the working memory.

Our results imply that one may not need overly complex neural network architectures and backpropagation through time to achieve a certain level of linguistic competence, encouraging more focus on simpler LMs in interdisciplinarily contextualizing the recent progress of neural LMs with the science of language.
As more specific future work, it might also be worthwhile to investigate various topologies in the recurrent architecture for future research, as discussed in previous ESN studies \citep{DengZ-2007-scale-free, KawaiPA-2019-small-world}.

%% file: section/02_esn.tex
\section{Echo State Network}
\label{section.esn}
\subsection{Definition and Initialization of ESN}
\label{section.definition-and-initialization-of-esn}

Let \( (\vu_{t} )_{t=1}^{T} \) be an input sequence, where \( \vu_{t} \in \R^{\vocabSize} \) (e.g., one-hot vector), and \(\vh_{0}=\bm{0}\).
ESN updates its state \(\vh_{t}\) and output \(\vo_{t}\) over time step \(t=0,\ldots,T\) as follows:
\begin{align}
\begin{split}
    \vh_{t+1}
    &=
    (\bm{1}-\vLeaking)\odot\vh_{t}
    \ifthenelse{\boolean{isdoublecolumn}}{\\&\quad}{}
    + \vLeaking\odot\hpActivation(\mRec\vh_{t}+\mInput\vu_{t+1}),
    \\
    \vo_{t+1} &= \mOutput\vh_{t+1}+\vBias,
\end{split}
\end{align}
where
\( \mInput \in \R^{\stateSize \times \vocabSize} \) is an input matrix,
\( \mRec \in \R^{\stateSize \times \stateSize} \) is a recurrent matrix,
\( \mOutput \in \R^{\vocabSize \times \stateSize} \) is an output matrix,
\( \vBias \in \R^{\vocabSize} \) is an output bias,
\(\vLeaking \in \R^{\stateSize} \) is a leaking rate,
\(\hpActivation\) is an element-wise activation function such as \(\tanh\),
and
\(\odot\) is the element-wise product.
\(\mOutput\) and \(\vBias\) are trainable, whereas \(\mInput\), \(\mRec\), and \(\vLeaking\) are frozen.

\paragraph{Initialization of \(\mInput\)}

First, we randomly generate matrices \(\mM_{\textup{in}}\) and \(\mV_{\textup{in}}\) of the same shape as the input matrix via element-wise i.i.d sampling:
\(
    (\mM_{\textup{in}})_{ij}
    \iidsample
    \mathrm{Bernoulli}(\hpConnectivity)
\)
and
\(
    (\mV_{\textup{in}})_{ij}
    \iidsample
    \mathcal{N}(0,\hpInputScale^{2}).
\)
Then, we obtain the input matrix as:
\begin{align}
\label{equation.init.input_matrix.2}
    \mInput=\mM_{\textup{in}}\odot\mV_{\textup{in}},
\end{align}
where \(\hpInputScale>0\) is an \emph{input scale} and
\(\hpConnectivity\in(0,1]\) is a \emph{connectivity}.
The benefit of introducing sparse connectivity to \(\mInput\) is discussed in \citet{Gallicchio-2020-sparsity}.

\paragraph{Initialization of \(\mRec\)}
Similarly to the input matrix initialization, we randomly generate matrices \(\mM_{\textup{rec}}\) and \(\mV_{\textup{rec}}\) as:
\(
    (\mM_{\textup{rec}})_{ij}
    \iidsample
    \mathrm{Bernoulli}(\hpConnectivity)
\) and
\(
    (\mV_{\textup{rec}})_{ij}
    \iidsample
    \mathcal{N}\left(0,1\right)
\).
Then, we obtain the recurrent matrix as follows:
\begin{align}
\label{equation.init.recurrent_matrix.2}
    \mRec
    =
    \frac{\hpSpectralRadius}{\rho(\mM_{\textup{rec}}\odot\mV_{\textup{rec}})}
    \mM_{\textup{rec}}\odot\mV_{\textup{rec}},
\end{align}
where \(\rho(\cdot)\) indicates the \emph{spectral radius} of a given matrix, and \(\hpSpectralRadius>0\) specifies the spectral radius of \(\mRec\).
In general, the spectral radius of a matrix \(\mM\) is defined as the largest value out of the absolute eigenvalues of \(\mM\), which intuitively indicates to what extent \(\mM\) can expand a vector \(\vv\) by \(\mM\vv\).
The ESN model's performance is empirically (and to some extent theoretically) known to be maximized in most cases if \(\hpSpectralRadius\) is sufficiently close to but never exceeds \(1\) \citep{Jaeger-2001-ESN}.
If \(\hpSpectralRadius\) is too small, the state \(\vh_{t}\) forgets the past input sequence too quickly.
if \( \hpSpectralRadius \) is too large, \( \vh_{t} \) evolves so chaotically that the generalization becomes almost impossible.

\paragraph{Initialization of \(\vLeaking\)}
The leaking rate vector \(\vLeaking\) is initialized as:
\begin{align}
    (\vLeaking)_{i} \iidsample \textrm{Uniform}(\hpMinLeakingRate,\hpMaxLeakingRate), \label{equation.init.leaking-rate}
\end{align}
where \(\hpMinLeakingRate\) and \(\hpMaxLeakingRate\) are hyperparameters specifying the minimum and maximum leaking rates respectively.
Although, in a typical formulation, the leaking rate is a scalar hyperparameter (or equivalently, just setting \(\hpMinLeakingRate=\hpMaxLeakingRate\)), the vectorized formulation is known to yield more complex, multi-scale dynamics \citep{TanakaMYA-2022-multiscale-reservoir}.
We expect it to give ESN a desirable language modeling property, recalling the gate mechanism's empirical success in LSTM \citep{HochreiterS-1997-lstm} and GRU \citep{ChoMBB-2014-gru}.

\paragraph{Parameterization of \(\mOutput\)}

The output matrix is usually parametrized as a full-rank dense matrix, but we parametrize it via low-rank decomposition to reduce the parameters, i.e., \( \mOutput = \mA\mB \), where \( \mA \in \R^{\vocabSize \times \hpOutputRank} \) and \( \mB \in \R^{\hpOutputRank \times \stateSize} \) are trainable matrices, with some \( \hpOutputRank < \min\{\stateSize, \vocabSize\} \).\footnote{
\(\mA\), \(\mB\), and \(\vBias\) are initialized with the PyTorch module (\texttt{nn.Linear}):
\((\mA)_{ij},~(\vBias)_{i}\iidsample\textrm{Uniform}(-\sqrt{\hpOutputRank^{-1}},\sqrt{\hpOutputRank^{-1}})\)
and
\((\mB)_{ij}\iidsample\textrm{Uniform}(-\sqrt{\stateSize^{-1}},\sqrt{\stateSize^{-1}})\).
}

We summarize ESN's hyperparameters in \cref{table.hyperparameters-of-esn} and parameters in \cref{table.parameters-of-esn}.

\begin{table}[tbp]
    \centering
    \ifthenelse{\boolean{isdoublecolumn}}{\small}{}
    \begin{tabular}{ll}
        \toprule
        \makecell{Hyperparameter name} &
        \makecell{Math symbol}
        \\
        \cmidrule(r){1-1}
        \cmidrule(l){2-2}
        state size &
        \( \stateSize \)
        \\
        spectral radius &
        \( \hpSpectralRadius \)
        \\
        input scale &
        \( \hpInputScale \)
        \\
        connectivity &
        \( \hpConnectivity~(=\hpRecDegree/\stateSize) \)
        \\
        minimum leaking rate &
        \( \hpMinLeakingRate \)
        \\
        maximum leaking rate &
        \( \hpMaxLeakingRate \)
        \\
        activation function &
        \( \hpActivation \)
        \\
        output rank &
        \( \hpOutputRank \)
        \\
        \bottomrule
    \end{tabular}
    \normalsize
    \ifthenelse{\boolean{isdoublecolumn}}{\vspace{-1.5ex}}{}
    \caption{Hyperparameters of ESN, adopted in this paper.}
    \ifthenelse{\boolean{isdoublecolumn}}{\vspace{-1.5ex}}{}
    \label{table.hyperparameters-of-esn}
\end{table}

\begin{table}[tbp]
    \centering
    \ifthenelse{\boolean{isdoublecolumn}}{\small}{}
    \begin{tabular}{llcl}
        \toprule
        \makecell{Parameter \\ name} &
        \makecell{Math \\ symbol} &
        \makecell{Frozen} &
        \makecell{Relevant hyper- \\ parameter(s)}
        \\
        \cmidrule(r){1-1}
        \cmidrule(rl){2-2}
        \cmidrule(rl){3-3}
        \cmidrule(l){4-4}
        input matrix &
        \(\mInput\) &
        \checkmark &
        \(\stateSize,\hpConnectivity,\hpInputScale\)
        \\
        recurrent matrix &
        \(\mRec\) &
        \checkmark &
        \(\stateSize,\hpConnectivity,\hpSpectralRadius\)
        \\
        leaking rate(s) &
        \(\vLeaking\) &
        \checkmark &
        \(\stateSize,\hpMinLeakingRate,\hpMaxLeakingRate\)
        \\
        output matrix &
        \(\mOutput\) &
        &
        \(\stateSize,\hpOutputRank\)
        \\
        output bias &
        \(\vBias\) &
        &
        \(\hpOutputRank\)
        \\
        \bottomrule
    \end{tabular}
    \normalsize
    \ifthenelse{\boolean{isdoublecolumn}}{\vspace{-1.5ex}}{}
    \caption{
        Parameters of ESN, adopted in this paper.
    }
    \ifthenelse{\boolean{isdoublecolumn}}{\vspace{-1.5ex}}{}
    \label{table.parameters-of-esn}
\end{table}

\subsection{ESN as Language Model}
\label{section.esn-as-language-model}

Let \( s = ( w_{t} )_{t=1}^{T} \) be an input sentence, where each \(w_{t} \in \{1,\ldots,\vocabSize\} \) indicates a token (word) index.
The first token \( w_{1} \) is the special symbol indicating the beginning-of-sentence (BOS), while the last token \( w_{T} \) is the special token indicating the end-of-sentence (EOS).
Following the standard practice in neural LMs, we obtain an input sequence \( ( \vu_{t} )_{t=1}^{T} \) by converting each index \(w_{t}\) into the corresponding one-hot vector \(\vu_{t} \in \R^{\vocabSize}\).
The model log-probability of \(s\) is given by:
\begin{align}
\begin{split}
    \ifthenelse{\boolean{isdoublecolumn}}{&}{}
    \log p(s \mid \trainableParam,\frozenParam)
    \ifthenelse{\boolean{isdoublecolumn}}{\\&}{}
    =
    \sum\nolimits_{t=1}^{T-1}
    \log p(w_{t+1} \mid (w_{i})_{i=1}^{t};\trainableParam,\frozenParam)
    \ifthenelse{\boolean{isdoublecolumn}}{\\&}{}
    =
    \sum\nolimits_{t=1}^{T-1}
    \log\left(\softmax(\vo_{t})\right)_{w_{t+1}},
\end{split}
\end{align}
where \( \trainableParam \) and \(\frozenParam\) are vectorizations of trainable and frozen parameters, respectively.

\subsection{Number of Parameters in ESN}
\label{section.number-of-parameters-of-esn}

In this section, we discuss the number of parameters in ESN.
We only count nonzero components as parameters and assume highly sparse connectivity, i.e., \( \hpConnectivity = \hpRecDegree / \stateSize \) for some \( \hpRecDegree \ll \stateSize \).
First, since \( \mInput \) is a \( \stateSize \times \vocabSize \)-matrix with connectivity \( \hpConnectivity \), we have
\begin{align}
    \#(\mInput)
    =
    \stateSize\vocabSize\hpConnectivity
    =
    \vocabSize \hpRecDegree,
\end{align}
where \( \#(\cdot) \) denotes the (expected) number of parameters.
Likewise, since \( \mRec \) is a \( \stateSize \times \stateSize \)-matrix with connectivity \( \hpConnectivity \), we have
\begin{align}
    \#(\mRec)
    =
    \stateSize^{2}\hpConnectivity
    =
    \stateSize \hpRecDegree.
\end{align}
As \( \mOutput \) is parametrized via low-rank decomposition, we have
\begin{align}
    \#(\mOutput) = (\vocabSize + \stateSize) \hpOutputRank.
\end{align}
Also, \( \#(\vLeaking) = \stateSize \) and \( \#(\vBias) = \vocabSize\).
To sum up, the numbers of frozen \(\frozenParam\), trainable \(\trainableParam\), and total parameters \(\totalParam\) are
\begin{align}
\begin{split}
    \#(\frozenParam)
    &= (\stateSize + \vocabSize) \hpRecDegree + \stateSize,
    \\
    \#(\trainableParam)
    &= (\stateSize + \vocabSize) \hpOutputRank + \vocabSize,
    \\
    \#(\totalParam)
    &= (\stateSize + \vocabSize)(\hpRecDegree + \hpOutputRank + 1).
\end{split}
\end{align}
Note that they grow only linearly with respect to \(\stateSize\), which enables efficient scaling.

%% file: section/03_why_esn.tex
\section{Why ESN?}
\label{section.why-esn}

In this section, we discuss again, but in more detail, why we should revisit ESN despite the era of LLMs.
As a matter of computational linguistics (CL), rather than engineering NLP directions, one would seek minimum conditions (aka. \textit{Occam's razor}) for some linguistic phenomena emerging.
In this sense, there is no necessity to begin with a complex neural network architecture like Transformer.
We should note that modern deep learning (e.g., Transformer) is a collection of heuristics whose counterparts in human brains are controversial; for example, humans do not seem (at least for now) to compute gradients by backpropagating the error, while deep learning history is considerably attributed to gradient stabilization, e.g., gradient clipping \citep{PascanuMB-2013-difficulty-rnn}, long-short term memory \citep{HochreiterS-1997-lstm}, Xavier/Kaiming initialization \citep{GlorotB-2010-Xavier-init, HeZRS-2015-Kaiming-init}, residual connection \citep{HeZRS-2016-resnet}, LayerNorm \citep{BaKH-2016-layernorm}, and learning rate scheduling \citep{GoyalDGNWKTJH-2018-accurate}.
Consequently, such complex designs may hinder the clear interpretation of its implication and what the model actually does in the study of CL, or at least we have to be aware of the potential risk of persisting with the particular architecture discovered by the narrow advancement of engineering-sided trial-and-errors, which may perhaps be too complex to progress the science of language.\footnote{%
Internal interpretation of Transformers, such as BERTology \citep[e.g.,][]{TenneyDP-2019-bert-pipeline} and mechanistic interpretability \citep[e.g.,][]{CunninghamERHS-2023-sparseautoencoder}, is of course a promising and intriguing direction, considering the recent success of LLMs.
}

ESN can, in contrast, be regarded as a natural discretization of a simple, continuous-time neuronal model \citep{Jaeger-2007-leaky}, which is suitable for exploring the minimum complexity.
Indeed, ESN is a linear regression model and thus free from backpropagation through time (BPTT) and layers.
More specifically, by regarding the update equation of the state as a feature function \(\bm{\phi}_{\frozenParam}(\cdot)\) from a (prefix of) input sequence \( (\vu_{t})_{t=1}^{T'} \) to a state \( \vh_{T'} \), one can restate ESN as a (log-)linear model
\(
    \vo_{T'} = \mOutput\bm{\phi}_{\frozenParam}\left( (\vu_{t})_{t=1}^{T'} \right) + \vBias
\).
Previous studies adopted ESN in the context of computational (psycho-)linguistics a decade ago \citep{TongBCC-2007-gammatical-esn, FrankC-2008-esn, FrankB-2011-insensitivity}, although they have now been replaced with modern deep learning methods \citep{GulordavaBGLB-2018-colorless-rnn, WilcoxGHQL-2020-predictive-power}.
Such a transition might be confounded by the advancement in optimization, scaling, or pre-processing techniques toward neural LMs rather than the essential differences in model architectures, and thus, it is worth revisiting ESN under modern settings. 
Notably, while the previous psycholinguistic works using ESN attempted at most \(\stateSize\approx 1000\), we scale it up to \(65,536\).\footnote{
PyTorch provides \texttt{torch.sparse} for efficient computation of sparse tensors.
This allows us to scale up ESN to some extent, even with a laboratory-level GPU resource.}

%% file: section/04_experiments.tex
\section{Experiments}
\label{section.experiments}

\begin{table*}[t]
    \newcommand{\cmdruleForMainResults}{
        \cmidrule(r){1-1}
        \cmidrule(rl){2-4}
        \cmidrule(rl){5-6}
        \cmidrule(l){7-7}
    }
    \centering
    \ifthenelse{\boolean{isdoublecolumn}}{\small}{}
    \begin{tabular}{
        l
        r
        r
        r
        r
        r
        r
    }
    \toprule
    {\makecell{LM name}} &
    {\makecell{\(\stateSize\)}} &
    {\makecell{\(\#(\trainableParam)\)\\{[M]}}} &
    {\makecell{\(\#(\totalParam)\)\\{[M]}}} &
    {\makecell{Train NLL\(\downarrow\)\\(per token)}} &
    {\makecell{Validation NLL\(\downarrow\)\\(per token)}} &
    {\makecell{BLiMP\(\uparrow\)\\{[\%]}}}
    \\
    \cmdruleForMainResults
    ESN            &  1,024 & 26 & 28 &
        \(5.311~(\pm 0.003)\) & \(5.024~(\pm 0.003)\) & \(56.2~(\pm 0.3)\) \\
    \hphantom{ESN} &  2,048 & 27 & 29 &
        \(5.173~(\pm 0.002)\) & \(4.887~(\pm 0.001)\) & \(56.1~(\pm 0.3)\) \\
    \hphantom{ESN} &  4,096 & 28 & 30 &
        \(5.071~(\pm 0.005)\) & \(4.794~(\pm 0.011)\) & \(57.9~(\pm 0.4)\) \\
    \hphantom{ESN} &  8,192 & 30 & 32 &
        \(4.992~(\pm 0.005)\) & \(4.708~(\pm 0.011)\) & \(58.5~(\pm 0.3)\) \\
    \hphantom{ESN} & 16,384 & 34 & 36 &
        \(4.965~(\pm 0.006)\) & \(4.642~(\pm 0.008)\) & \(59.2~(\pm 0.4)\) \\
    \hphantom{ESN} & 32,768 & 43 & 45 &
        \(4.987~(\pm 0.023)\) & \(4.630~(\pm 0.008)\) & \(60.0~(\pm 0.3)\) \\
    \hphantom{ESN} & 65,536 & 59 & 63 &
        \(5.102~(\pm 0.023)\) & \(4.690~(\pm 0.023)\) & \(60.5~(\pm 0.2)\) \\
    \cmdruleForMainResults
    GPT2 Scratch   &    768 & 124 & 124 &
        \(5.667~(\pm 0.024)\) & \(4.803~(\pm 0.011)\) & \(58.7~(\pm 0.7)\) \\
    LSTM           &    512 &  54 &  54 &
        \(4.503~(\pm 0.000)\) & \(4.120~(\pm 0.001)\) & \(67.8~(\pm 0.2)\) \\
    \cmdruleForMainResults
    GPT2 OpenAI    &    768 & 124 & 124 &
        {-} & {-} & 82.2 \\
    \bottomrule
    \end{tabular}
    \vspace{-1ex}
    \caption{
        The number of trainable parameters \(\#(\trainableParam)\), the total number of parameters \(\#(\totalParam)\), the train NLL (per token), the validation NLL (per token), and the overall BLiMP score [\%].
        The train NLLs are higher than the validation NLLs because the former are averaged over batches during training, while the latter are computed after one epoch.
        \( (\pm~\cdot~) \) represents one standard error of mean, computed from 4 runs for each configuration.
    }
    \vspace{-2ex}
    \label{table.results}
\end{table*}
\subsection{Experimental Setup}

\paragraph{Dataset and Preprocessing}

We used the BabyLM Challenge dataset (2023 version) \citep{WarstadtCMWWZ-2023-babylm, ChoshenCHLMRWWWZ-2024-babylm} as the training and validation datasets.
The training data consists of approximately 100M words.
Since they were provided as raw texts, we applied the NLTK sentence tokenizer\footnote{\url{https://www.nltk.org/api/nltk.tokenize.sent_tokenize.html}} and then applied the GPT2 tokenizer\footnote{ \url{https://huggingface.co/docs/transformers/model_doc/gpt2\#transformers.GPT2Tokenizer}}.
Thus, \( \vocabSize = 50257 \).
The maximum sequence length per sentence was set to 512, and sequences shorter than 6 were removed so that LMs would learn longer dependency structures.\footnote{Minimum sentence length should be 4 due to BOS/EOS.}

\paragraph{ESN's Configuration}

We set
\( \hpSpectralRadius = 0.99 \),
\( \hpInputScale = 1 \),
\( \hpRecDegree = 32 \),
\( \hpMinLeakingRate = 0 \),
\( \hpMaxLeakingRate = 1 \),
\( \hpActivation(\cdot) = \tanh(\cdot) \),
and
\( \hpOutputRank = 512 \).
We varied the state size as \( \stateSize \in \{ 2^{10}, 2^{11}, 2^{12}, 2^{13}, 2^{14}, 2^{15}, 2^{16} \} \), according to which \(\hpConnectivity=\hpRecDegree/\stateSize\) also changed.\footnote{
We also investigated the influence of the connectivity and leaking rates.
See Appendices \ref{section.influence-of-sparse-connectivity} and \ref{section.influence-of-leaking-rates} respectively.
}

\paragraph{Comparison Targets}

We compare ESNs with the GPT2 pre-trained by OpenAI (GPT2 OpenAI), a GPT2 trained from scratch (GPT2 Scratch), and an LSTM with an embedding and hidden state of size 512.
GPT2 Scratch followed the default configuration of \texttt{GPT2Config}.\footnote{\url{https://huggingface.co/docs/transformers/model_doc/gpt2\#transformers.GPT2Config}}
We applied the dropout (\(p=0.1\)) immediately after the embedding and before the output matrix application in LSTM.

\paragraph{Training Procedure}

The LMs were trained for just 1 epoch.
The batch size was 32.
AdamW \citep{LoshchilovH-2019-adamw} was used as an optimizer with its default parameters.\footnote{\url{https://pytorch.org/docs/stable/generated/torch.optim.AdamW.html}}

\paragraph{Syntactic Evaluation}

To measure the extent to which the LMs accurately capture syntactic structure, we used BLiMP \citep{WarstadtPLMPWB-2020-blimp}, a dataset of minimal pairs of English sentences covering various syntactic phenomena.

\subsection{Results}

Table \ref{table.results} shows the number of training parameters \(\#(\trainableParam)\), the total number of parameters \(\#(\totalParam)\), the train negative log-likelihood (NLL), the validation NLL, and the overall BLiMP score.
Somewhat surprisingly, ESN outperforms GPT2 Scratch in the validation NLL when \(\stateSize\geq 4096\) and in the BLiMP score when \(\stateSize\geq 16384\), even though ESN has less trainable/total parameters than GPT2.
The scaling generally improved the performance of ESN, like the scaling law \citep{KaplanMHBCCGRWA-2020-neural-scaling-law}, monotonically for BLiMP scores, while the train and validation NLL improved until \(\stateSize = 16384\); we suspect this is just due to the limited training time (i.e., 1 epoch).
These tentatively suggest that large ESNs can also learn core syntactic phenomena.
GPT2 OpenAI achieves the highest BLiMP score, which is not surprising for its pre-training scale, while this is still useful to grasp the ``upper bound'' of LMs at a similar scale.
Excluding GPT2 OpenAI, LSTM exhibits the best results in the train NLL, validation NLL, and BLiMP; the inferiority of the Transformer to LSTM might reflect the transformer's data inefficiency in the BabyLM setting. 

%% file: section/05_conclusion.tex
\section{Conclusion}
\label{section.conclusion}

This paper briefly revisited a basic reservoir computing model called Echo State Network (ESN) as a neural language model.
The experimental results showed that ESN is comparable to or superior to the Transformer model, at least on the BabyLM scale.
The best result of LSTM suggests that it is worth revisiting the benefit of gate mechanisms as well.
Exploring better topologies in the ESN's architecture will also be worthwhile beyond the simple, sparse connectivity adopted in this paper.

%% file: appendix/a_connectivity.tex
\section{Influence of Sparse Connectivity}
\label{section.influence-of-sparse-connectivity}

\begin{table*}[h]
    \newcommand{\cmdruleInThisTable}{
        \cmidrule(r){1-1}
        \cmidrule(rl){2-5}
        \cmidrule(rl){6-7}
        \cmidrule(l){8-8}
    }
    \centering
    \begin{tabular}{
        l
        r
        l
        r
        r
        r
        r
        r
    }
    \toprule
    {\makecell{LM name}} &
    {\makecell{\(\stateSize\)}} &
    {\makecell{\(\hpConnectivity\)}} &
    {\makecell{\(\#(\trainableParam)\)\\{[M]}}} &
    {\makecell{\(\#(\totalParam)\)\\{[M]}}} &
    {\makecell{Train NLL\(\downarrow\)\\(per token)}} &
    {\makecell{Validation NLL\(\downarrow\)\\(per token)}} &
    {\makecell{BLiMP\(\uparrow\)\\{[\%]}}}
    \\
    \cmdruleInThisTable
    ESN &  4,096 & \(2^{-12}\) & 28 & 28 &
        \(5.792~(\pm 0.027)\) & \(5.497~(\pm 0.027)\) & \(55.3~(\pm 0.7)\)
        \\
    \hphantom{ESN} & & \(2^{-11}\) & & 28 &
        \(5.387~(\pm 0.036)\) & \(5.080~(\pm 0.037)\) & \(56.1~(\pm 0.4)\)
       \\
    \hphantom{ESN} & & \(2^{-10}\) & & 28 &
        \(5.137~(\pm 0.018)\) & \(4.833~(\pm 0.017)\) & \(56.8~(\pm 0.6)\)
        \\
    \hphantom{ESN} & & \(2^{-9}\) & & 28 &
        \(5.063~(\pm 0.005)\) & \(4.761~(\pm 0.002)\) & \(58.3~(\pm 0.3)\)
        \\
    \hphantom{ESN} & & \(2^{-8}\) & & 29 &
        \(5.055~(\pm 0.006)\) & \(4.756~(\pm 0.002)\) & \(56.9~(\pm 0.3)\)
        \\
    \hphantom{ESN} & & \(2^{-7}\) & & 30 &
        \(5.071~(\pm 0.005)\) & \(4.794~(\pm 0.011)\) & \(57.9~(\pm 0.4)\) \\
    \hphantom{ESN} & & \(2^{-6}\) & & 31 &
        \(5.094~(\pm 0.001)\) & \(4.820~(\pm 0.017)\) & \(57.6~(\pm 0.5)\) \\
    \hphantom{ESN} & & \(2^{-5}\) & & 35 &
        \(5.141~(\pm 0.003)\) & \(4.836~(\pm 0.005)\) & \(56.7~(\pm 0.1)\) \\
    \hphantom{ESN} & & \(2^{-4}\) & & 42 &
        \(5.178~(\pm 0.004)\) & \(4.890~(\pm 0.014)\) & \(57.7~(\pm 0.3)\) \\
    \hphantom{ESN} & & \(2^{-3}\) & & 56 &
        \(5.228~(\pm 0.001)\) & \(4.943~(\pm 0.011)\) & \(56.9~(\pm 0.5)\) \\
    \hphantom{ESN} & & \(2^{-2}\) & & 84 &
        \(5.286~(\pm 0.001)\) & \(4.994~(\pm 0.015)\) & \(56.6~(\pm 0.3)\) \\
    \hphantom{ESN} & & \(2^{-1}\) & & 139 &
        \(5.350~(\pm 0.001)\) & \(5.041~(\pm 0.008)\) & \(56.7~(\pm 0.2)\) \\
    \hphantom{ESN} & & \(1\) & & 251 &
        \(5.417~(\pm 0.001)\) & \(5.102~(\pm 0.008)\) & \(56.3~(\pm 0.3)\) \\
    \bottomrule
    \end{tabular}
    \caption{
        The number of trainable parameters \(\#(\trainableParam)\), the total number of parameters \(\#(\totalParam)\), the train NLL (per token), the validation NLL (per token), and the overall BLiMP score [\%], when the connectivity \(\hpConnectivity\) is varied.
        The train NLLs are higher than the validation NLLs because the former are averaged over batches during training, while the latter are computed after one epoch.
        \( (\pm~\cdot~) \) represents one standard error of mean, computed from 4 runs for each configuration.
    }
    \label{table.connectivity-results}
\end{table*}
We conducted an additional experiment to investigate the influence of sparse connectivity, as the high sparsity \(\hpConnectivity\ll 1\) adopted in this paper is not a common practice in deep learning-based language modeling.
We set
\(\hpSpectralRadius=0.99\),
\(\hpInputScale=1\),
\(\hpMinLeakingRate=0\),
\(\hpMaxLeakingRate=1\),
\(\hpActivation(\cdot)=\tanh(\cdot)\),
and
\(\hpOutputRank=512\),
following the configuration of the main experiment.
In this section, however, we fixed \(\stateSize=2^{12}=4096\) while varied \(\hpRecDegree\) as \(
    \hpRecDegree
    \in
    \{1, 2, 2^{2}, 2^{3}, 2^{4}, 2^{5}(=32), 2^{6}, 2^{7}, 2^{8}, 2^{9}, 2^{10}, 2^{11}, 2^{12} \}
\), i.e., \(
    \hpConnectivity
    \in
    \{ 2^{-12}, 2^{-11}, 2^{-10}, 2^{-9}, \allowbreak 2^{-8}, \allowbreak 2^{-7}, \allowbreak 2^{-6}, 2^{-5}, \allowbreak 2^{-4}, \allowbreak 2^{-3}, 2^{-2}, 2^{-1}, 1\}
\).
The results are shown in Table \ref{table.connectivity-results}.
More sparse connectivity results in better validation NLL score until \(\hpConnectivity=2^{-8}\), whereas further sparsity degrades the score.
The BliMP scores show a similar tendency.
They suggest surprisingly that the (appropriate) sparse connectivity benefits not only the efficient scaling but also the capability of language.

%% file: appendix/b_leaking_rate.tex
\section{Influence of Leaking Rates}
\label{section.influence-of-leaking-rates}

\begin{table*}[h]
    \newcommand{\cmdruleInThisTable}{
        \cmidrule(r){1-1}
        \cmidrule(rl){2-5}
        \cmidrule(rl){6-7}
        \cmidrule(l){8-8}
    }
    \centering
    \begin{tabular}{
        l
        r
        l
        r
        r
        r
        r
        r
    }
    \toprule
    {\makecell{LM name}} &
    {\makecell{\(\stateSize\)}} &
    {\makecell{\(\hpMinLeakingRate\)}} &
    {\makecell{\(\#(\trainableParam)\)\\{[M]}}} &
    {\makecell{\(\#(\totalParam)\)\\{[M]}}} &
    {\makecell{Train NLL\(\downarrow\)\\(per token)}} &
    {\makecell{Validation NLL\(\downarrow\)\\(per token)}} &
    {\makecell{BLiMP\(\uparrow\)\\{[\%]}}}
    \\
    \cmdruleInThisTable
    ESN &  4,096 & \(0\) & 28 & 30 &
        \(5.071~(\pm 0.005)\) & \(4.794~(\pm 0.011)\) & \(57.9~(\pm 0.4)\) \\
    \hphantom{ESN} & & \(0.1\) & & &
        \(5.093~(\pm 0.007)\) & \(4.803~(\pm 0.014)\) & \(57.7~(\pm 0.3)\) \\
    \hphantom{ESN} & & \(0.2\) & & &
        \(5.105~(\pm 0.006)\) & \(4.823~(\pm 0.015)\) & \(57.6~(\pm 0.5)\) \\
    \hphantom{ESN} & & \(0.3\) & & &
        \(5.120~(\pm 0.005)\) & \(4.829~(\pm 0.009)\) & \(57.8~(\pm 0.4)\) \\
    \hphantom{ESN} & & \(0.4\) & & &
        \(5.136~(\pm 0.004)\) & \(4.845~(\pm 0.007)\) & \(57.7~(\pm 0.4)\) \\
    \hphantom{ESN} & & \(0.5\) & & &
        \(5.155~(\pm 0.004)\) & \(4.861~(\pm 0.006)\) & \(57.5~(\pm 0.4)\) \\
    \hphantom{ESN} & & \(0.6\) & & &
        \(5.177~(\pm 0.004)\) & \(4.884~(\pm 0.006)\) & \(57.4~(\pm 0.4)\) \\
    \hphantom{ESN} & & \(0.7\) & & &
        \(5.202~(\pm 0.004)\) & \(4.909~(\pm 0.006)\) & \(57.4~(\pm 0.4)\) \\
    \hphantom{ESN} & & \(0.8\) & & &
        \(5.233~(\pm 0.004)\) & \(4.936~(\pm 0.005)\) & \(57.2~(\pm 0.4)\) \\
    \hphantom{ESN} & & \(0.9\) & & &
        \(5.272~(\pm 0.005)\) & \(4.971~(\pm 0.004)\) & \(57.1~(\pm 0.3)\) \\
    \hphantom{ESN} & & \(1\) & & &
        \(5.330~(\pm 0.006)\) & \(5.018~(\pm 0.005)\) & \(56.8~(\pm 0.2)\) \\
    \bottomrule
    \end{tabular}
    \caption{
        The number of trainable parameters \(\#(\trainableParam)\), the total number of parameters \(\#(\totalParam)\), the train NLL (per token), the validation NLL (per token), and the overall BLiMP score [\%], when the minimum leaking rate \(\hpMinLeakingRate\) is varied.
        The train NLLs are higher than the validation NLLs because the former are averaged over batches during training, while the latter are computed after one epoch.
        \( (\pm~\cdot~) \) represents one standard error of mean, computed from 4 runs for each configuration.
    }
    \label{table.leaking-rate-results}
\end{table*}

We conducted another experiment to investigate the influence of leaking rates.
We set
\(\hpSpectralRadius=0.99\),
\(\hpInputScale=1\),
\(\hpRecDegree=32\),
\(\hpMaxLeakingRate=1\),
\(\hpActivation(\cdot)=\tanh(\cdot)\),
and
\(\hpOutputRank=512\),
following the configuration of the main experiment.
In this section, however, we fixed \(\stateSize=2^{12}=4096\) while varied \(\hpMinLeakingRate\) as \(\hpMinLeakingRate\in\{0,0.1,0.2,0.3,0.4,0.5,0.6,0.7,0.8,0.9,1\}\).
The results are shown in Table \ref{table.leaking-rate-results}.
Lower \(\hpMinLeakingRate\) results in lower NLLs and higher BLiMP scores, which suggests that the multiple time scales indeed benefit the capability of language.

%% file: main_arxiv.bbl
\begin{thebibliography}{33}
\providecommand{\natexlab}[1]{#1}
\providecommand{\url}[1]{\texttt{#1}}
\expandafter\ifx\csname urlstyle\endcsname\relax
  \providecommand{\doi}[1]{doi: #1}\else
  \providecommand{\doi}{doi: \begingroup \urlstyle{rm}\Url}\fi

\bibitem[Ba et~al.(2016)Ba, Kiros, and Hinton]{BaKH-2016-layernorm}
Jimmy~Lei Ba, Jamie~Ryan Kiros, and Geoffrey~E. Hinton.
\newblock Layer normalization, 2016.
\newblock URL \url{https://arxiv.org/abs/1607.06450}.

\bibitem[Cabessa et~al.(2021)Cabessa, Hernault, Kim, Lamonato, and Levy]{CabessaHKLL-2021-efficient-text-classification}
Jérémie Cabessa, Hugo Hernault, Heechang Kim, Yves Lamonato, and Yariv~Z. Levy.
\newblock Efficient text classification with echo state networks.
\newblock In \emph{2021 International Joint Conference on Neural Networks (IJCNN)}, pages 1--8, 2021.
\newblock \doi{10.1109/IJCNN52387.2021.9533958}.

\bibitem[Cho et~al.(2014)Cho, van Merrienboer, Bahdanau, and Bengio]{ChoMBB-2014-gru}
Kyunghyun Cho, Bart van Merrienboer, Dzmitry Bahdanau, and Yoshua Bengio.
\newblock On the properties of neural machine translation: Encoder-decoder approaches.
\newblock In Dekai Wu, Marine Carpuat, Xavier Carreras, and Eva~Maria Vecchi, editors, \emph{Proceedings of SSST@EMNLP 2014, Eighth Workshop on Syntax, Semantics and Structure in Statistical Translation, Doha, Qatar, 25 October 2014}, pages 103--111. Association for Computational Linguistics, 2014.
\newblock \doi{10.3115/V1/W14-4012}.
\newblock URL \url{https://aclanthology.org/W14-4012/}.

\bibitem[Choshen et~al.(2024)Choshen, Cotterell, Hu, Linzen, Mueller, Ross, Warstadt, Wilcox, Williams, and Zhuang]{ChoshenCHLMRWWWZ-2024-babylm}
Leshem Choshen, Ryan Cotterell, Michael~Y. Hu, Tal Linzen, Aaron Mueller, Candace Ross, Alex Warstadt, Ethan Wilcox, Adina Williams, and Chengxu Zhuang.
\newblock [call for papers] the 2nd babylm challenge: Sample-efficient pretraining on a developmentally plausible corpus.
\newblock \emph{CoRR}, abs/2404.06214, 2024.
\newblock \doi{10.48550/ARXIV.2404.06214}.
\newblock URL \url{https://doi.org/10.48550/arXiv.2404.06214}.

\bibitem[Cunningham et~al.(2023)Cunningham, Ewart, Riggs, Huben, and Sharkey]{CunninghamERHS-2023-sparseautoencoder}
Hoagy Cunningham, Aidan Ewart, Logan Riggs, Robert Huben, and Lee Sharkey.
\newblock Sparse autoencoders find highly interpretable features in language models, 2023.
\newblock URL \url{https://arxiv.org/abs/2309.08600}.

\bibitem[Deng and Zhang(2007)]{DengZ-2007-scale-free}
Zhidong Deng and Yi~Zhang.
\newblock Collective behavior of a small-world recurrent neural system with scale-free distribution.
\newblock \emph{{IEEE} Trans. Neural Networks}, 18\penalty0 (5):\penalty0 1364--1375, 2007.
\newblock \doi{10.1109/TNN.2007.894082}.
\newblock URL \url{https://doi.org/10.1109/TNN.2007.894082}.

\bibitem[Frank and Bod(2011)]{FrankB-2011-insensitivity}
Stefan~L. Frank and Rens Bod.
\newblock Insensitivity of the human sentence-processing system to hierarchical structure.
\newblock \emph{Psychological Science}, 22\penalty0 (6):\penalty0 829--834, 2011.
\newblock \doi{10.1177/0956797611409589}.
\newblock URL \url{https://doi.org/10.1177/0956797611409589}.
\newblock PMID: 21586764.

\bibitem[Frank and Čerňanský(2008)]{FrankC-2008-esn}
Stefan~L. Frank and Michal Čerňanský.
\newblock Generalization and systematicity in echo state networks.
\newblock \emph{Proceedings of the Annual Meeting of the Cognitive Science Society}, 30\penalty0 (30), 2008.

\bibitem[Gallicchio(2020)]{Gallicchio-2020-sparsity}
Claudio Gallicchio.
\newblock Sparsity in reservoir computing neural networks, 2020.
\newblock URL \url{https://arxiv.org/abs/2006.02957}.

\bibitem[Glorot and Bengio(2010)]{GlorotB-2010-Xavier-init}
Xavier Glorot and Yoshua Bengio.
\newblock Understanding the difficulty of training deep feedforward neural networks.
\newblock In Yee~Whye Teh and D.~Mike Titterington, editors, \emph{Proceedings of the Thirteenth International Conference on Artificial Intelligence and Statistics, {AISTATS} 2010, Chia Laguna Resort, Sardinia, Italy, May 13-15, 2010}, volume~9 of \emph{{JMLR} Proceedings}, pages 249--256. JMLR.org, 2010.
\newblock URL \url{http://proceedings.mlr.press/v9/glorot10a.html}.

\bibitem[Goyal et~al.(2018)Goyal, Dollár, Girshick, Noordhuis, Wesolowski, Kyrola, Tulloch, Jia, and He]{GoyalDGNWKTJH-2018-accurate}
Priya Goyal, Piotr Dollár, Ross Girshick, Pieter Noordhuis, Lukasz Wesolowski, Aapo Kyrola, Andrew Tulloch, Yangqing Jia, and Kaiming He.
\newblock Accurate, large minibatch sgd: Training imagenet in 1 hour, 2018.
\newblock URL \url{https://arxiv.org/abs/1706.02677}.

\bibitem[Gulordava et~al.(2018)Gulordava, Bojanowski, Grave, Linzen, and Baroni]{GulordavaBGLB-2018-colorless-rnn}
Kristina Gulordava, Piotr Bojanowski, Edouard Grave, Tal Linzen, and Marco Baroni.
\newblock Colorless green recurrent networks dream hierarchically.
\newblock In \emph{Proceedings of the 2018 Conference of the North American Chapter of the Association for Computational Linguistics: Human Language Technologies, {NAACL-HLT} 2018, New Orleans, Louisiana, USA, June 1-6, 2018, Volume 1 (Long Papers)}, pages 1195--1205. Association for Computational Linguistics, 2018.
\newblock \doi{10.18653/V1/N18-1108}.
\newblock URL \url{https://doi.org/10.18653/v1/n18-1108}.

\bibitem[He et~al.(2015)He, Zhang, Ren, and Sun]{HeZRS-2015-Kaiming-init}
Kaiming He, Xiangyu Zhang, Shaoqing Ren, and Jian Sun.
\newblock Delving deep into rectifiers: Surpassing human-level performance on imagenet classification.
\newblock In \emph{2015 {IEEE} International Conference on Computer Vision, {ICCV} 2015, Santiago, Chile, December 7-13, 2015}, pages 1026--1034. {IEEE} Computer Society, 2015.
\newblock \doi{10.1109/ICCV.2015.123}.
\newblock URL \url{https://doi.org/10.1109/ICCV.2015.123}.

\bibitem[He et~al.(2016)He, Zhang, Ren, and Sun]{HeZRS-2016-resnet}
Kaiming He, Xiangyu Zhang, Shaoqing Ren, and Jian Sun.
\newblock Deep residual learning for image recognition.
\newblock In \emph{2016 {IEEE} Conference on Computer Vision and Pattern Recognition, {CVPR} 2016, Las Vegas, NV, USA, June 27-30, 2016}, pages 770--778. {IEEE} Computer Society, 2016.
\newblock \doi{10.1109/CVPR.2016.90}.
\newblock URL \url{https://doi.org/10.1109/CVPR.2016.90}.

\bibitem[Hochreiter and Schmidhuber(1997)]{HochreiterS-1997-lstm}
Sepp Hochreiter and J{\"{u}}rgen Schmidhuber.
\newblock Long short-term memory.
\newblock \emph{Neural Comput.}, 9\penalty0 (8):\penalty0 1735--1780, 1997.
\newblock \doi{10.1162/neco.1997.9.8.1735}.
\newblock URL \url{https://doi.org/10.1162/neco.1997.9.8.1735}.

\bibitem[Jaeger(2001)]{Jaeger-2001-ESN}
Herbert Jaeger.
\newblock {The ``echo state'' approach to analysing and training recurrent neural networks}.
\newblock Technical report, German National Research Center for Information Technology GMD Technical Report 148, 2001.
\newblock Erratum note available at \url{https://www.ai.rug.nl/minds/uploads/EchoStatesTechRepErratum.pdf}.

\bibitem[Jaeger et~al.(2007)Jaeger, Lukoševičius, Popovici, and Siewert]{Jaeger-2007-leaky}
Herbert Jaeger, Mantas Lukoševičius, Dan Popovici, and Udo Siewert.
\newblock Optimization and applications of echo state networks with leaky-integrator neurons.
\newblock \emph{Neural Networks}, 20\penalty0 (3):\penalty0 335--352, 2007.
\newblock ISSN 0893-6080.
\newblock \doi{https://doi.org/10.1016/j.neunet.2007.04.016}.
\newblock URL \url{https://www.sciencedirect.com/science/article/pii/S089360800700041X}.
\newblock Echo State Networks and Liquid State Machines.

\bibitem[Kaplan et~al.(2020)Kaplan, McCandlish, Henighan, Brown, Chess, Child, Gray, Radford, Wu, and Amodei]{KaplanMHBCCGRWA-2020-neural-scaling-law}
Jared Kaplan, Sam McCandlish, Tom Henighan, Tom~B. Brown, Benjamin Chess, Rewon Child, Scott Gray, Alec Radford, Jeffrey Wu, and Dario Amodei.
\newblock Scaling laws for neural language models, 2020.
\newblock URL \url{https://arxiv.org/abs/2001.08361}.

\bibitem[Kawai et~al.(2019)Kawai, Park, and Asada]{KawaiPA-2019-small-world}
Yuji Kawai, Jihoon Park, and Minoru Asada.
\newblock A small-world topology enhances the echo state property and signal propagation in reservoir computing.
\newblock \emph{Neural Networks}, 112:\penalty0 15--23, 2019.
\newblock \doi{10.1016/J.NEUNET.2019.01.002}.
\newblock URL \url{https://doi.org/10.1016/j.neunet.2019.01.002}.

\bibitem[Loshchilov and Hutter(2019)]{LoshchilovH-2019-adamw}
Ilya Loshchilov and Frank Hutter.
\newblock Decoupled weight decay regularization.
\newblock In \emph{7th International Conference on Learning Representations, {ICLR} 2019, New Orleans, LA, USA, May 6-9, 2019}. OpenReview.net, 2019.
\newblock URL \url{https://openreview.net/forum?id=Bkg6RiCqY7}.

\bibitem[Luko{\v{s}}evi{\v{c}}ius(2012)]{Lukosevicius-2012-practical}
Mantas Luko{\v{s}}evi{\v{c}}ius.
\newblock \emph{A Practical Guide to Applying Echo State Networks}, pages 659--686.
\newblock Springer Berlin Heidelberg, Berlin, Heidelberg, 2012.
\newblock ISBN 978-3-642-35289-8.
\newblock \doi{10.1007/978-3-642-35289-8_36}.
\newblock URL \url{https://doi.org/10.1007/978-3-642-35289-8_36}.

\bibitem[Lukoševičius and Jaeger(2009)]{LukoseviciusJ-2009-reservoir}
Mantas Lukoševičius and Herbert Jaeger.
\newblock Reservoir computing approaches to recurrent neural network training.
\newblock \emph{Computer Science Review}, 3\penalty0 (3):\penalty0 127--149, 2009.
\newblock ISSN 1574-0137.
\newblock \doi{https://doi.org/10.1016/j.cosrev.2009.03.005}.
\newblock URL \url{https://www.sciencedirect.com/science/article/pii/S1574013709000173}.

\bibitem[Pascanu et~al.(2013)Pascanu, Mikolov, and Bengio]{PascanuMB-2013-difficulty-rnn}
Razvan Pascanu, Tomas Mikolov, and Yoshua Bengio.
\newblock On the difficulty of training recurrent neural networks, 2013.
\newblock URL \url{https://arxiv.org/abs/1211.5063}.

\bibitem[Rodan and Tino(2011)]{RodanT-2011-minimum-complexity-esn}
Ali Rodan and Peter Tino.
\newblock Minimum complexity echo state network.
\newblock \emph{IEEE Transactions on Neural Networks}, 22\penalty0 (1):\penalty0 131--144, 2011.
\newblock \doi{10.1109/TNN.2010.2089641}.

\bibitem[Tanaka et~al.(2019)Tanaka, Yamane, H{\'{e}}roux, Nakane, Kanazawa, Takeda, Numata, Nakano, and Hirose]{TanakaYHNKTNNH-2019-recent}
Gouhei Tanaka, Toshiyuki Yamane, Jean~Benoit H{\'{e}}roux, Ryosho Nakane, Naoki Kanazawa, Seiji Takeda, Hidetoshi Numata, Daiju Nakano, and Akira Hirose.
\newblock Recent advances in physical reservoir computing: {A} review.
\newblock \emph{Neural Networks}, 115:\penalty0 100--123, 2019.
\newblock \doi{10.1016/J.NEUNET.2019.03.005}.
\newblock URL \url{https://doi.org/10.1016/j.neunet.2019.03.005}.

\bibitem[Tanaka et~al.(2022)Tanaka, Matsumori, Yoshida, and Aihara]{TanakaMYA-2022-multiscale-reservoir}
Gouhei Tanaka, Tadayoshi Matsumori, Hiroaki Yoshida, and Kazuyuki Aihara.
\newblock Reservoir computing with diverse timescales for prediction of multiscale dynamics.
\newblock \emph{Phys. Rev. Res.}, 4:\penalty0 L032014, Jul 2022.
\newblock \doi{10.1103/PhysRevResearch.4.L032014}.
\newblock URL \url{https://link.aps.org/doi/10.1103/PhysRevResearch.4.L032014}.

\bibitem[Tenney et~al.(2019)Tenney, Das, and Pavlick]{TenneyDP-2019-bert-pipeline}
Ian Tenney, Dipanjan Das, and Ellie Pavlick.
\newblock {BERT} rediscovers the classical {NLP} pipeline.
\newblock In \emph{Proceedings of the 57th Conference of the Association for Computational Linguistics, {ACL} 2019, Florence, Italy, July 28- August 2, 2019, Volume 1: Long Papers}, pages 4593--4601. Association for Computational Linguistics, 2019.
\newblock \doi{10.18653/V1/P19-1452}.
\newblock URL \url{https://doi.org/10.18653/v1/p19-1452}.

\bibitem[Tong et~al.(2007)Tong, Bickett, Christiansen, and Cottrell]{TongBCC-2007-gammatical-esn}
Matthew~H. Tong, Adam~D. Bickett, Eric~M. Christiansen, and Garrison~W. Cottrell.
\newblock Learning grammatical structure with echo state networks.
\newblock \emph{Neural Networks}, 20\penalty0 (3):\penalty0 424--432, 2007.
\newblock \doi{10.1016/J.NEUNET.2007.04.013}.
\newblock URL \url{https://doi.org/10.1016/j.neunet.2007.04.013}.

\bibitem[Vaswani et~al.(2017)Vaswani, Shazeer, Parmar, Uszkoreit, Jones, Gomez, Kaiser, and Polosukhin]{VaswaniSPUJGKP-2017-transformer}
Ashish Vaswani, Noam Shazeer, Niki Parmar, Jakob Uszkoreit, Llion Jones, Aidan~N Gomez, \L~ukasz Kaiser, and Illia Polosukhin.
\newblock Attention is all you need.
\newblock In \emph{Advances in Neural Information Processing Systems}, volume~30. Curran Associates, Inc., 2017.
\newblock URL \url{https://proceedings.neurips.cc/paper_files/paper/2017/file/3f5ee243547dee91fbd053c1c4a845aa-Paper.pdf}.

\bibitem[Warstadt et~al.(2020)Warstadt, Parrish, Liu, Mohananey, Peng, Wang, and Bowman]{WarstadtPLMPWB-2020-blimp}
Alex Warstadt, Alicia Parrish, Haokun Liu, Anhad Mohananey, Wei Peng, Sheng{-}Fu Wang, and Samuel~R. Bowman.
\newblock Blimp: The benchmark of linguistic minimal pairs for english.
\newblock \emph{Trans. Assoc. Comput. Linguistics}, 8:\penalty0 377--392, 2020.
\newblock \doi{10.1162/TACL\_A\_00321}.
\newblock URL \url{https://doi.org/10.1162/tacl\_a\_00321}.

\bibitem[Warstadt et~al.(2023)Warstadt, Choshen, Mueller, Williams, Wilcox, and Zhuang]{WarstadtCMWWZ-2023-babylm}
Alex Warstadt, Leshem Choshen, Aaron Mueller, Adina Williams, Ethan Wilcox, and Chengxu Zhuang.
\newblock Call for papers - the babylm challenge: Sample-efficient pretraining on a developmentally plausible corpus.
\newblock \emph{CoRR}, abs/2301.11796, 2023.
\newblock \doi{10.48550/ARXIV.2301.11796}.
\newblock URL \url{https://doi.org/10.48550/arXiv.2301.11796}.

\bibitem[Wieting and Kiela(2019)]{WietingK-2019-no}
John Wieting and Douwe Kiela.
\newblock No training required: Exploring random encoders for sentence classification.
\newblock In \emph{7th International Conference on Learning Representations, {ICLR} 2019, New Orleans, LA, USA, May 6-9, 2019}. OpenReview.net, 2019.
\newblock URL \url{https://openreview.net/forum?id=BkgPajAcY7}.

\bibitem[Wilcox et~al.(2020)Wilcox, Gauthier, Hu, Qian, and Levy]{WilcoxGHQL-2020-predictive-power}
Ethan Wilcox, Jon Gauthier, Jennifer Hu, Peng Qian, and Roger Levy.
\newblock On the predictive power of neural language models for human real-time comprehension behavior.
\newblock In \emph{Proceedings of the 42th Annual Meeting of the Cognitive Science Society - Developing a Mind: Learning in Humans, Animals, and Machines, CogSci 2020, virtual, July 29 - August 1, 2020}. cognitivesciencesociety.org, 2020.
\newblock URL \url{https://cogsci.mindmodeling.org/2020/papers/0375/index.html}.

\end{thebibliography}
